\documentclass[english]{cccconf}
\usepackage[comma,numbers,square,sort&compress]{natbib}
\usepackage{verbatim}
\usepackage{epstopdf}
\usepackage{amsmath,amssymb,amsfonts}
\usepackage{algorithm}
\usepackage{algorithmic}
\usepackage{graphicx}
\usepackage{textcomp}
\usepackage{xcolor}
\usepackage{subfigure}
\usepackage{booktabs}
\usepackage{caption}
\usepackage{multirow}

\usepackage{mathrsfs}

\usepackage{booktabs}
\usepackage{varwidth}
\usepackage{graphicx}

\usepackage{tabularx}

\usepackage[figuresright]{rotating}

\usepackage[colorlinks,
linkcolor=blue,
anchorcolor=blue,
citecolor=blue
]{hyperref}
\UseRawInputEncoding

\begin{document}

\title{Tuning-Free Structured Sparse PCA via Deep \\Unfolding Networks}

\author{Long Chen and Xianchao Xiu}

\affiliation[]{School of Mechatronic Engineering and Automation, Shanghai University,
        Shanghai 200444, China
        \email{150538cli@shu.edu.cn, xcxiu@shu.edu.cn}}
\maketitle

\begin{abstract}
Sparse principal component analysis (PCA) is a well-established dimensionality reduction technique that is often used for unsupervised feature selection (UFS). However, determining the regularization parameters is rather challenging, and conventional approaches, including grid search and Bayesian optimization, not only bring great computational costs but also exhibit high sensitivity.
To address these limitations, we first establish a structured sparse PCA formulation by integrating  $\ell_1$-norm and $\ell_{2,1}$-norm to capture the local and global structures, respectively. Building upon the off-the-shelf alternating direction method of multipliers (ADMM) optimization framework, we then design an interpretable deep unfolding network that translates iterative optimization steps into trainable neural architectures. This innovation enables automatic learning of the regularization parameters, effectively bypassing the empirical tuning requirements of conventional methods. Numerical experiments on benchmark datasets validate the advantages of our proposed method over the existing state-of-the-art methods. Our code will be accessible at \href{https://github.com/xianchaoxiu/SPCA-Net}{https://github.com/xianchaoxiu/SPCA-Net}.
\end{abstract}


\footnotetext{This work was supported by the National Natural Science Foundation of China under Grant 12371306. (\textit{Corresponding author: Xianchao Xiu}.)}

\section{Introduction}
Unsupervised feature selection (UFS) has emerged as a critical technique in high-dimensional data analysis, particularly for image and signal processing applications where class labels are often unavailable; see \cite{li2017feature,solorio2020review,li2024exploring} for a recent survey. Among various approaches, sparse principal component analysis (PCA) \cite{zou2018selective} has become a prominent UFS tool, demonstrating remarkable success across diverse domains ranging from  image processing \cite{rekavandi2024learning,zhou2024unsupervised}, fault diagnosis \cite{xiu2020laplacian,xiu2022sparsity} to natural language processing \cite{drikvandi2023sparse,jahan2023systematic}.

Let $A\in \mathbb{R}^{d\times n}$ denote the data matrix. Mathematically, the classical PCA can be formulated as
\begin{equation}\label{pca}
	\begin{aligned}
		\min_{X } \quad & \frac{1}{2} \| A - XX^{\top} A \|^{2}_{\textrm{F}}  \\
		\textrm{s.t.}\quad  &X^\top X=I, 
	\end{aligned}
\end{equation}
where $X\in \mathbb{R}^{d\times m}$ is the projection matrix and $I\in \mathbb{R}^{m\times m}$ is the identity matrix. However, it suffers from limited feature interpretability. To address this issue, sparse PCA was proposed in \cite{zou2006sparse} by defining
\begin{equation}\label{CSPCA}
	\begin{aligned}
		\min_{X } \quad & \frac{1}{2} \| A - XX^{\top} A \|^{2}_{\textrm{F}} +\lambda \|X\|_{1}\\
		\textrm{s.t.}\quad &X^\top X=I, 
	\end{aligned}
\end{equation}
where $\|X\|_1$ is the $\ell_{1}$-norm, defined as the sum of absolute values ​​of all elements, and $\lambda$ is the regularization parameter.
By enforcing the $\ell_{1}$-norm and adjusting $\lambda$, sparse PCA can obtain a sparse  projection matrix, thereby enhancing the interpretability and alleviating the interference of noise.  However, two fundamental challenges remain.

On the one hand, the $\ell_1$-norm in sparse PCA only exploits element-wise sparsity, which often lead to biased solutions \cite{xiu2020data}. This limitation becomes particularly apparent in applications where features naturally exhibit group-wise sparsity, which cannot be characterized by $\ell_1$-norm but could be captured by $\ell_{2,1}$-norm  (defined as the sum of $\ell_2$-norm of each row) \cite{liu2009multi,nie2010efficient}. Current methodologies usually consider these sparsity-inducing norms separately or apply them to different variables, while it remains an open question whether the integration of $\ell_1$-norm and $\ell_{2,1}$-norm with PCA could yield superior performance in the field of UFS.

On the other hand, traditional parameter selection strategies (e.g., grid search, Bayesian optimization) face significant scalability limitations in high-dimensional settings \cite{snoek2012practical,shang2021ell,tian2022comprehensive}.  The emerging paradigm of deep unfolding networks \cite{gregor2010learning} offers a promising alternative by embedding optimization mechanics into neural architectures through algorithmic unfolding, i.e., transforming the iterative algorithm steps into network layers \cite{monga2021algorithm,zhang2023physics,shlezinger2023model}. This approach combines mathematical interpretability with neural network efficiency, achieving excellent performance in many fields such as communications \cite{nguyen2024joint}, compressive sensing \cite{yang2018admm}, and infrared small target detection \cite{wu2024rpcanet}. Notably, deep unfolding networks typically require fewer trainable parameters than conventional deep architectures, which substantially reduces the computational complexity and training costs \cite{mukherjee2023learned,chen2024learning}. Despite these advancements, the application of deep unfolding networks to PCA-based UFS remains absent from the literature.

\begin{figure*}[t]
	\centering
	\includegraphics[width=1.0\textwidth]{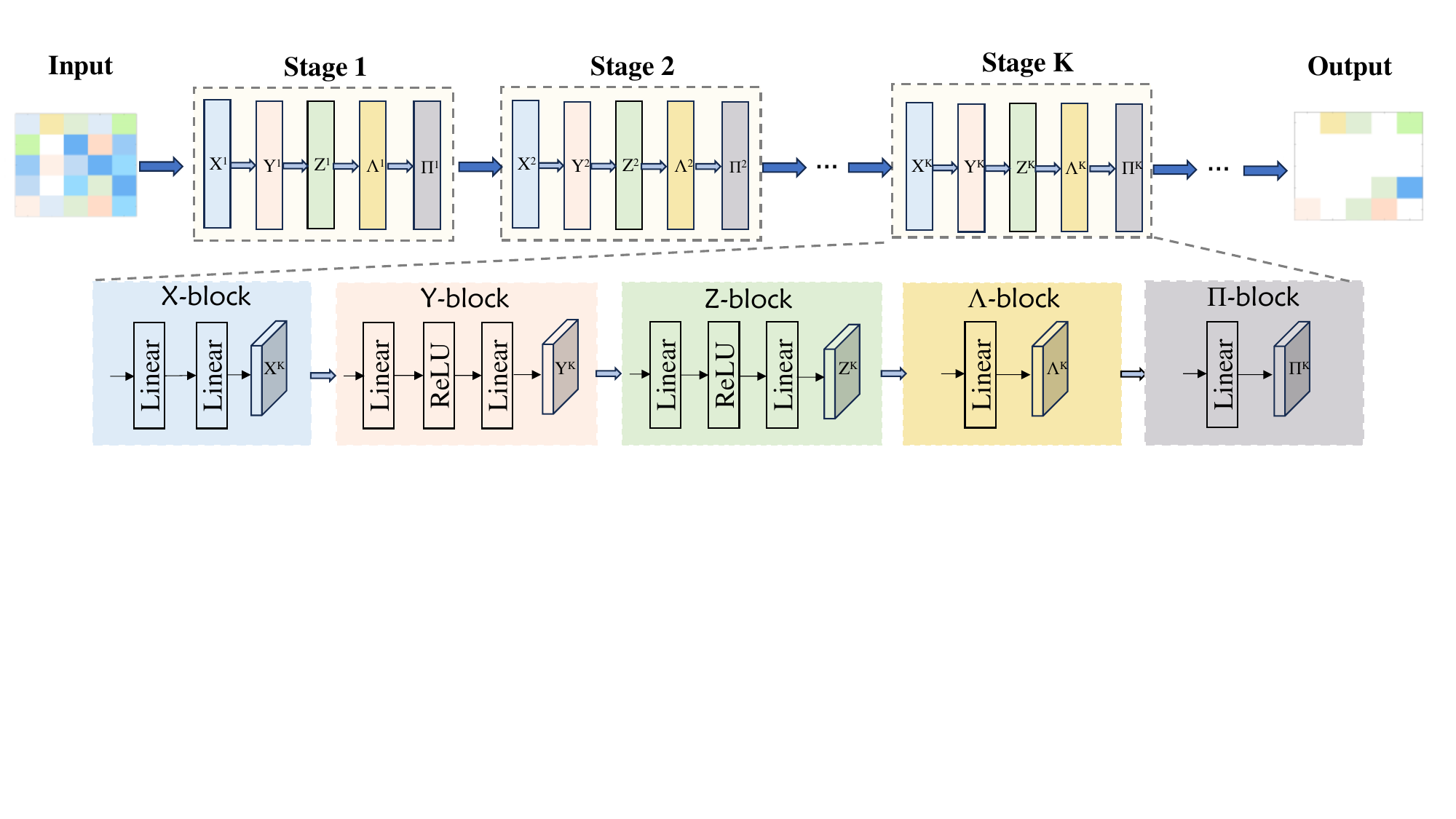}
	\vspace*{-43mm}
	\caption{Overview of the proposed SPCA-Net.}
	\label{pcanet}
\end{figure*}

Inspired by sparse group Lasso \cite{simon2013sparse}, we propose the following structured sparse PCA formulation
\begin{equation}\label{USPCA}
	\begin{aligned} 
		\min_{X}  \quad  &\frac{1}{2} \| A - XX^{\top} A \|^{2}_{\textrm{F}} + \lambda \| X \|_{2,1} + \mu \| X \|_1 \\
		\textrm{s.t.}  \quad&X^\top X = I,
	\end{aligned}
\end{equation}
where $ \lambda, \mu >0$ are the regularization parameters, $\ell_1$-norm and $\ell_{2,1}$-norm are introduced to capture local and global structures, respectively, thus play a complementary role. Departing from conventional first-order optimization algorithms with manual parameter tuning, we develop an intelligent solution through deep unfolding networks that automatically learns optimal parameter configurations. For convenience, we call this method SPCA-Net.

The contributions of this paper are twofold.
\begin{itemize}
	\item A novel structured sparse PCA framework integrating $\ell_1$-norm  and $\ell_{2,1}$-norm is proposed to enhance UFS.
	\item A pioneering deep unfolding networks-based  algorithm is developed that combines mathematical priors with data-driven efficiency; see Fig. \ref{pcanet}.
\end{itemize}

\section{Methodology}
This section describes how to leverage deep unfolding networks to develop a tuning-free algorithm for problem \eqref{USPCA}. Technically, we unfold each step of  the alternating direction method of multipliers (ADMM) \cite{han2022survey} into a layer in neural networks, thus transforming parameters into learnable ones.

We first introduce auxiliary variables $Y, Z\in \mathbb{R}^{d\times m}$ and reformulate problem \eqref{USPCA} as
\begin{equation}\label{problem}
    \begin{aligned}
    \min_{X, Y, Z} \quad & \frac{1}{2} \|A - X X^{\top} A\|^{2}_{\textrm{F}} + \lambda \|Y\|_{2,1} + \mu \|Z\|_1 \\
    \textrm{s.t.}  \quad~&X^{\top} X = I,~ X =Y,~X = Z.
\end{aligned}
\end{equation}
The corresponding augmented Lagrangian function is
\begin{equation}
    \begin{aligned}
  &  \mathcal{L}(X,Y,Z,\Lambda,\Pi)\\
   &=  \frac{1}{2} \|A - X X^{\top} A\|^{2}_{\textrm{F}} + \lambda \|Y\|_{2,1}+\mu \|Z\|_1 \\ 
    & ~~~+ \langle \Lambda, X - Y \rangle+ \frac{\alpha}{2} \|X - Y\|^{2}_{\textrm{F}} \\
    &~~~+ \langle \Pi, X - Z \rangle+ \frac{\beta}{2} \|X - Z\|^{2}_{\textrm{F}},
    \end{aligned}
\end{equation}
where $\Lambda$ and $\Pi$ are the Lagrange multipliers, and $\alpha, \beta > 0$ are 
penalty parameters. Now we can update one variable while fixing others, which is the well-known ADMM.

\subsection{Update $X^{k+1}$}
It can be equivalently transformed into 
\begin{equation}\label{x1}
    \begin{aligned}
    \min_{X} \quad & \frac{1}{2} \|A - X X^{\top} A\|^{2}_{\textrm{F}}+\frac{\alpha}{2} \|X - Y^k + \Lambda^k / \alpha\|^{2}_{\textrm{F}}\\
    & + \frac{\beta}{2} \|X - Z^k + \Pi^k / \beta\|^{2}_{\textrm{F}} \\
    \textrm{s.t.}\quad & X^{\top} X = I,
\end{aligned}
\end{equation}
which does not have a closed-form solution. Denote the objective function of \eqref{x1} as $f(X)$. Using linearization with approximation parameter $\eta>0$, we derive
\begin{equation}
    \begin{aligned}
    \min_{X} \quad & f(X^k) + \langle \nabla f(X^k), X - X^k \rangle\\
                   &  + \frac{1}{2\eta} \|X - X^k\|^{2}_{\textrm{F}} \\
    \textrm{s.t.} \quad & X^{\top} X = I,
\end{aligned}
\end{equation}
which can be simplified to
\begin{equation}\label{pp-1}
\begin{aligned}
    \min_{X} \quad& ( AA^{\top} X^k +  ( \alpha (X^k - Y^k  + \Lambda^k  / \alpha)\\
    &+ \beta (X^k - Z^k  + \Pi^k  / \beta)) X + \frac{1}{2\eta} \|X - X^k\|^{2}_{\textrm{F}} \\
    \textrm{s.t.} \quad & X^{\top} X = I.
\end{aligned}
\end{equation}
Denote $ M^k=AA^{\top} X^k+( \alpha (X^k - Y^k  + {\Lambda^k }/{\alpha}) + \beta (X^k - Z^k  + {\Pi^k }/{\beta})  )$, then problem \eqref{pp-1} can be rewritten as
\begin{equation}
    \begin{aligned}
    \min_X \quad & \frac{1}{2\eta} \| X - X^k + \eta M^k \|^{2}_{\textrm{F}} \\
    \textrm{s.t.} \quad & X^{\top} X = I.
\end{aligned}
\end{equation}
Let $X^k - \eta M^k = U\Sigma V^\top$ be the SVD. Then
\begin{equation}\label{uv}
    X^{k+1} = U V^\top,
\end{equation}
which can be described by the following network
\begin{equation}\label{xnet}
      X^{k+1} =\textbf{\textit{LargNet}}(U, V^\top),
\end{equation}
where \textbf{\textit{LargNet}} denotes  two distinct linear layers, as shown in Fig. \ref{pcanet}.

\subsection{Update $Y^{k+1}$}
Once  $X$ has been updated,  $Y$ can be obtained by solving
\begin{equation}\label{y1}
    \min_Y~  \lambda \| Y \|_{2,1} + \frac{\alpha}{2} \| X^{k+1}  - Y + {\Lambda^k}/{\alpha} \|^{2}_{\textrm{F}}.
\end{equation}
According to the  soft thresholding associated with $\ell_{2,1}$-norm and \cite{li2023lrr}, it has the close-form solution
\begin{equation}
	\frac{X^{k+1}+ \Lambda^k / \alpha}{\| X^{k+1} + \Lambda^k / \alpha \|_2}  \text{ReLU}(\| X^{k+1}+ \Lambda^k / \alpha \|_2- \lambda / \alpha),
\end{equation}
which can be denoted by the following network
\begin{equation}\label{ynet}
    Y^{k+1} = \textbf{\textit{GSoftNet}} (X^{k+1}+ \Lambda^k / \alpha, \lambda/\alpha).
\end{equation}
We would like to emphasize that this paper uses \textbf{\textit{GSoftNet}} to denote networks involving $\ell_{2,1}$-norm, and the word \textbf{G} comes from group sparsity.

\begin{algorithm}[t]
	\caption{SPCA-Net}\label{net}
	\textbf{Input:} Data $A \in \mathbb{R}^{d \times n}$, parameters $\lambda, \mu, \alpha, \beta$ \\
	\textbf{Initialize:} ($X^0$, $Y^0$, $Z^0$, $\Lambda^0$, $\Pi^0$) \\
	\textbf{While} not converged \textbf{do}
	\begin{algorithmic}[1]
		\STATE Update $X^{k+1}$ by 
		\begin{equation}\nonumber
			X^{k+1} =\textbf{\textit{LargNet}}(U, V^\top)
		\end{equation}
		\STATE Update $Y^{k+1}$ by 
		\begin{equation}\nonumber
		Y^{k+1}=	\textbf{\textit{GSoftNet}} (X^{k+1}+ \Lambda^k / \alpha, \lambda/\alpha)
		\end{equation}
		\STATE Update $Z^{k+1}$ by 
		\begin{equation}\nonumber
			Z^{k+1} =\textbf{\textit{SoftNet}}(X^{k+1} + \Pi^k / \beta, \mu/\beta)
		\end{equation}
		\STATE Update $\Lambda^{k+1}, \Pi^{k+1}$ by 
		\begin{equation}\nonumber
			\begin{aligned}
				\Lambda^{k+1}&=\textbf{\textit{Linear}}(\Lambda^{k}, X^{k+1},Y^{k+1}, \alpha)\\
				\Pi^{k+1}&=\textbf{\textit{Linear}}(\Pi^{k}, X^{k+1}, Z^{k+1}, \beta)
			\end{aligned}
		\end{equation}
	\end{algorithmic}
	\textbf{End while}\\
	\textbf{Output:} Trained $X$
\end{algorithm}

\subsection{Update $Z^{k+1}$}
After $X, Y$ have been updated, $Z$ can be solved by 
\begin{equation}\label{z1}
    \min_Z ~ \mu \| Z \|_1 + \frac{\beta}{2} \| X^{k+1} - Z + {\Pi^k}/{\beta} \|^{2}_{\textrm{F}}.
\end{equation}
Similar to the $Y$-subproblem, invoking the  soft thresholding associated with $\ell_{1}$-norm, it has the close-form solution
\begin{equation}
    \frac{X^{k+1}+ \Pi^k / \beta}{| X^{k+1} + \Pi^k / \beta |}  \text{ReLU}(| X^{k+1}+ \Pi^k / \beta | - \mu/ \beta),
\end{equation}
which can be characterized by the following network
\begin{equation}\label{znet}
    Z^{k+1} =\textbf{\textit{SoftNet}}(X^{k+1} + \Pi^k / \beta, \mu/\beta).
\end{equation}
Here, \textbf{\textit{SoftNet}} denotes networks that involve $\ell_{1}$-norm.

\subsection{Update $\Lambda^{k+1},\Pi^{k+1}$}
The Lagrange multipliers can be calculated by 
\begin{equation}\label{ww}
	\begin{aligned}
	\Lambda^{k+1}&=\Lambda^{k}+ \alpha (X^{k+1} - Y^{k+1}),\\
	\Pi^{k+1}&=\Pi^{k}+ \beta (X^{k+1} - Z^{k+1}),
	\end{aligned}
\end{equation}
which can be represented by the following networks
\begin{equation}
    \begin{aligned}
    \Lambda^{k+1}&=\textbf{\textit{Linear}}(\Lambda^{k}, X^{k+1}, Y^{k+1}, \alpha),\\
    \Pi^{k+1}&=\textbf{\textit{Linear}}(\Pi^{k}, X^{k+1}, Z^{k+1}, \beta).
    \end{aligned}
\end{equation}

In summary, the complete deep unfolding architecture for solving problem \eqref{USPCA} is presented in Algorithm \ref{net} and Fig. \ref{pcanet}. Our framework eliminates manual parameter tuning through full parameter learnability, encompassing both regularization papameters $(\lambda, \mu)$ and penalty parameters $(\alpha, \beta)$.  In addition, the loss function is defined as
\begin{equation}
	\textrm{Loss}=  \frac{1}{2}\| A- \bar{X}  \bar{X}^{\top} A \|_\textrm{F}^2  + \lambda \| \bar{X} \|_{2,1} + \mu  \| \bar{X} \|_1,
\end{equation}
where $\bar{X}$ represents the predicted matrix.

\section{Experiments}

 \begin{table}[t] \label{data}
	\caption{The dataset information.}\label{data}
	\centering
	\setlength{\tabcolsep}{1.5mm}{
		\begin{tabular}{|c|c|c|c|c|}
			\hline
			~Datasets~ & ~Features~ & ~Samples~ & ~Categories~ & ~Types~   \\
			\hline\hline
			COIL20  & 1024 & 1440  & 20 &  Image   \\ \hline
			 Isolet & 617 & 1560  & 26  & Speech  \\ \hline
			 UMIST &  644  &  575  & 20 &  Handwritten \\ \hline
	  	MSTAR  &1024 & 2425  & 10  & Biological  \\ 
			\hline
		\end{tabular}
	}
\end{table}

\begin{figure*}[t]
	\vspace*{-37mm}
	\centering
	\subfigcapskip=-113pt
	\subfigure[COIL20]{
		\label{c}
		\centering
		\includegraphics[width=0.5\textwidth]{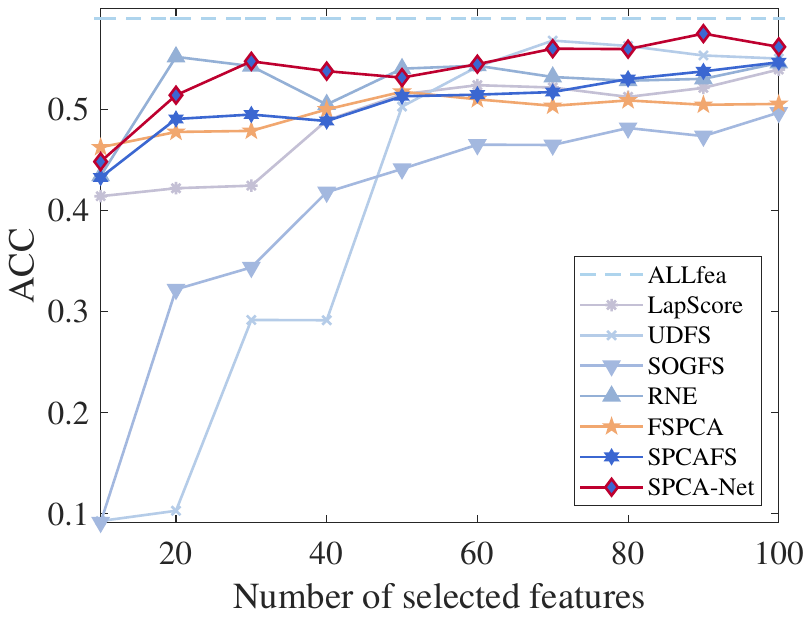}
	}\hspace{-25mm} 
	\subfigcapskip=-113pt
	\vspace*{-73mm}
	\subfigure[Isolet]{
		\label{d}
		\centering
		\includegraphics[width=0.5\textwidth]{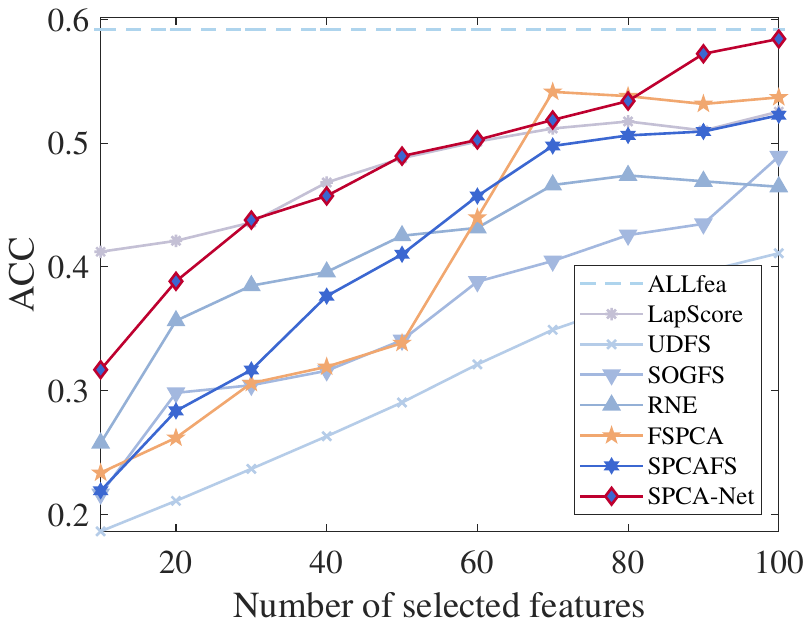} 
	}
	\subfigcapskip=-113pt
	\subfigure[UMIST]{
		\label{c}
		\centering
		\includegraphics[width=0.5\textwidth]{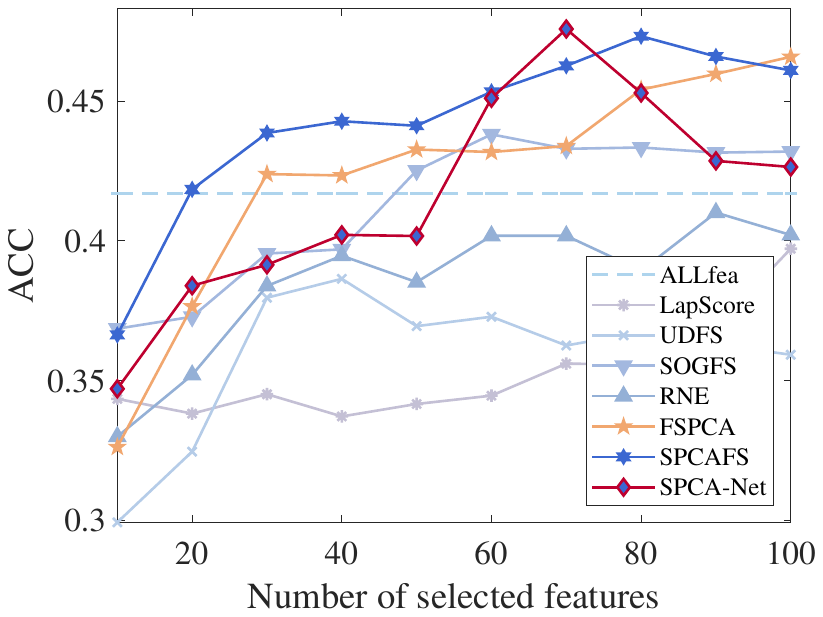}
	}\hspace{-25mm} 
	\subfigcapskip=-113pt
	\subfigure[MSTAR]{
		\label{d}
		\centering
		\includegraphics[width=0.5\textwidth]{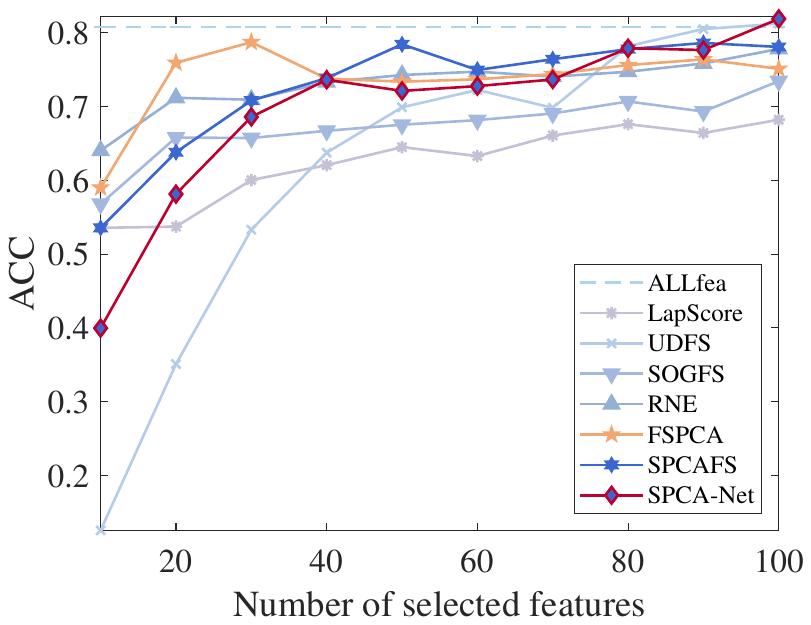}
	}
	\vspace*{-34mm}
	\caption{Visualization of ACC results for all compared methods.}
	\label{pcanetacc}
\end{figure*}

\begin{figure*}[t]
	\vspace*{-37mm}
	\centering
	\subfigcapskip=-113pt
	\subfigure[COIL20]{
		\label{c}
		\centering
		\includegraphics[width=0.5\textwidth]{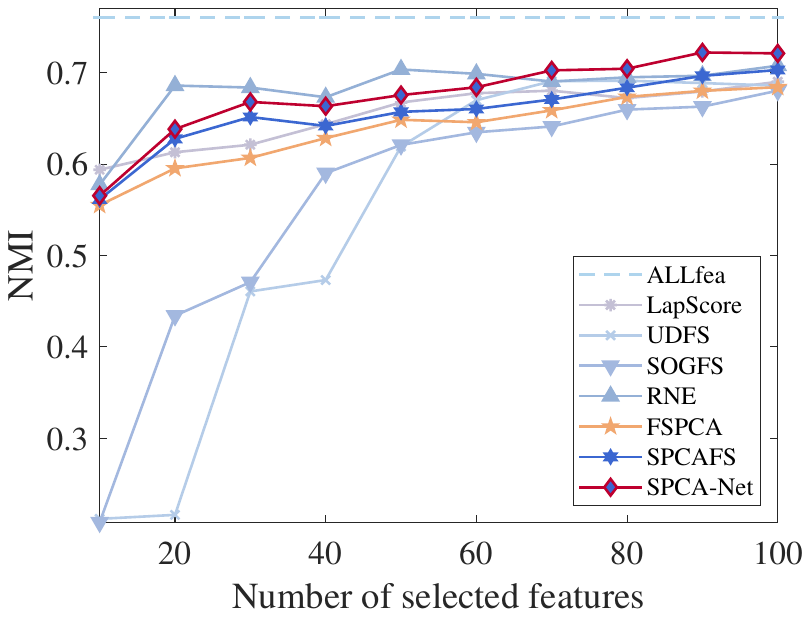}
	}\hspace{-25mm} 
	\subfigcapskip=-113pt
	\vspace*{-73mm}
	\subfigure[Isolet]{
		\label{d}
		\centering
		\includegraphics[width=0.5\textwidth]{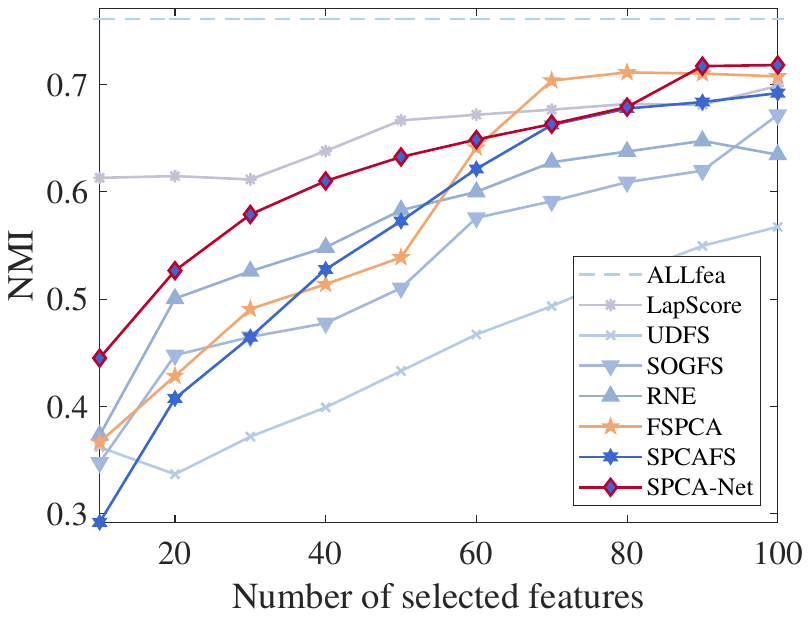} 
	}
	\subfigcapskip=-113pt
	\subfigure[UMIST]{
		\label{c}
		\centering
		\includegraphics[width=0.5\textwidth]{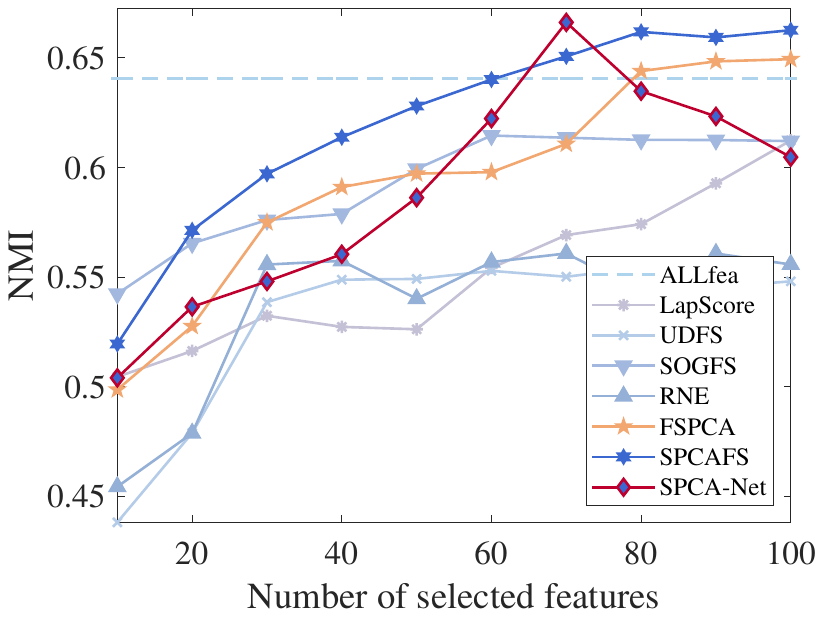}
	}\hspace{-25mm} 
	\subfigcapskip=-113pt
	\subfigure[MSTAR]{
		\label{d}
		\centering
		\includegraphics[width=0.5\textwidth]{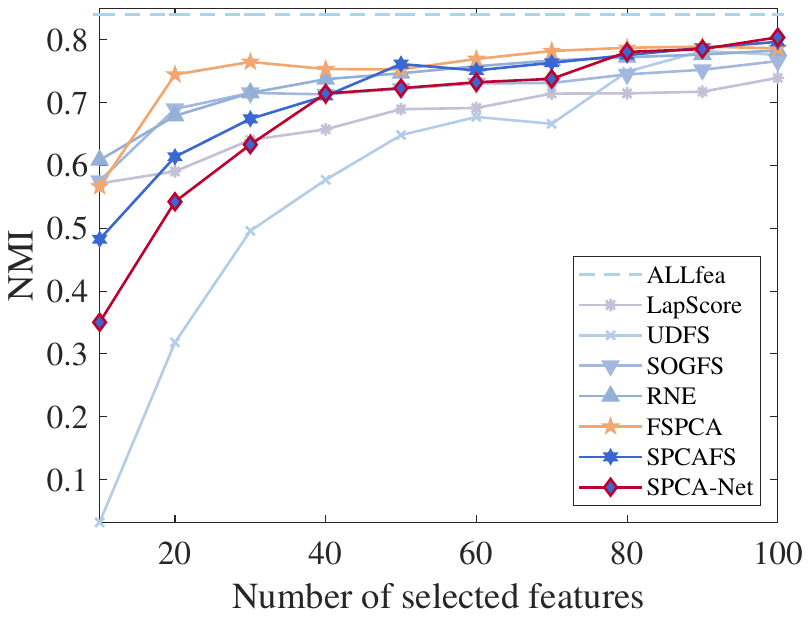}
	}
	\vspace*{-34mm}
	\caption{Visualization of NMI results for all compared methods.}
	\label{plot-acc}
\end{figure*}

\begin{table*}[t]
	\renewcommand\arraystretch{1.15}
	\caption{ACC results (mean \% $\pm$ std \%) for all compared methods.} \label{nmi}
	\centering%
	\setlength{\tabcolsep}{1.5mm}{
		\begin{tabular}{|c|c|c|c|c|c|c|c|c|c|c|}%
			\hline
		~Datasets~   & ALLfea & LapScore  \cite{he2005laplacian} & UDFS  \cite{yang2011ℓ} & SOGFS  \cite{nie2016unsupervised} & RNE \cite{liu2020robust} & FSPCA  \cite{nie2022learning} & SPCAFS  \cite{li2021sparse} & SPCA-Net  \\ \hline\hline
			\multirow{2}{*}{COIL20} & 58.97$\pm$4.99 & 53.91$\pm$3.61 & 56.70$\pm$3.09 & 49.66$\pm$3.63 & 55.16$\pm$3.35 & 51.71$\pm$3.05 & 54.63$\pm$3.64 & \textbf{57.46$\pm$2.76} \\
			& (10) & (100) & (70) & (100) & (20) & (50) & (100) & \textbf{(90)} \\  \hline
			\multirow{2}{*}{Isolet} & 59.18$\pm$3.19 & 52.55$\pm$2.83 & 41.11$\pm$1.71 & 48.93$\pm$2.69 & 47.39$\pm$2.91 & 54.15$\pm$2.69 & 52.26$\pm$2.81 & \textbf{58.43$\pm$4.31} \\ 
			& (10) & (100) & (100) & (100) & (80) & (70) & (100) & \textbf{(100)} \\ \hline
			\multirow{2}{*}{UMIST} & 41.68$\pm$2.46 & 39.71$\pm$3.28 & 38.64$\pm$1.61 & 43.81$\pm$2.98 & 41.01$\pm$2.25 & 46.58$\pm$2.34 & 47.32$\pm$3.48 & \textbf{47.58$\pm$4.97} \\
			& (10) & (100) & (40) & (80) & (90) & (100) & (80) & \textbf{(70)} \\  \hline
			\multirow{2}{*}{MSTAR} & 80.81$\pm$8.76 & 68.21$\pm$4.57 & 81.25$\pm$7.48 & 73.46$\pm$5.61 & 77.82$\pm$6.16 & 78.74$\pm$5.20 & 78.63$\pm$8.68 & \textbf{81.90$\pm$6.87} \\ 
			& (10) & (100) & (100) & (100) & (100) & (30) & (90) & \textbf{(100)} \\ 
			\hline
		\end{tabular}
	}\label{pcanet_acc}
\end{table*}

 This section demonstrates  the superiority of our proposed method for the UFS task on four public datasets; see Table \ref{data} for more details.

 \subsection{Parameter Setup}
 In experiments, the compared methods include LapScore \cite{he2005laplacian}, UDFS \cite{yang2011ℓ}, SOGFS \cite{nie2016unsupervised}, RNE \cite{liu2020robust}, FSPCA \cite{nie2022learning}, and SPCAFS \cite{li2021sparse}.  Among them, LapScore, UDFS, SOGFS, and RNE are directly implemented by the AutoUFSTool toolbox\footnote{https://github.com/farhadabedinzadeh/AutoUFSTool}. SPCAFS can be downloaded from the author's link\footnote{https://github.com/quiter2005/algorithm}. For the above methods, the default parameters are adopted.
 For our proposed SPCA-Net, the number of network layers is set to 5. In addition,
all features marked as ALLfea are used as baselines for comparison.
 
To evaluate the compared methods, two popular metrics, i.e., clustering accuracy (ACC) and normalized mutual information (NMI), are used. As suggested in \cite{li2021sparse}, the number of selected features is chosen from 10 to 100. To ensure fairness and reduce the variations caused by different initial conditions, the $k$-means clustering algorithm is repeated 50 times. Therefore, the final results are presented as the average and standard deviation of these 50 runs.

 \subsection{Numerical Results}
 Fig. \ref{pcanetacc} and Fig. \ref{nmi} show the comparative ACC and NMI metrics across feature dimensions, respectively.  It can be seen that as the number of selected features increases, the metrics of all compared methods improve to some extent, but in most cases, our proposed SPCA-Net can achieve the best performance. Although our proposed SPCA-Net does not perform particularly well when there are fewer selected features, as the number  increases, the performance will improve significantly until it exceeds others.

 Further, Table \ref{pcanet_acc} and Table \ref{pcanet_nmi} quantify the corresponding ACC and NMI results, respectively. Additionally, their average values and standard deviations are provided, with the best results marked in bold. It can be concluded that for all datasets, our proposed SPCA-Net consistently outperforms the other compared methods.  In particular, for the Isolet dataset, the ACC value of SPCA-Net is 59.17\%, and the corresponding NMI value is 75.35\%, which are 5.02\% and 4.20\% higher than the second-place method, respectively.

\begin{table*}[t]\label{pcanet_nmi}
	\renewcommand\arraystretch{1.15}
	\caption{NMI results (mean  \% $\pm$ std  \%) for all compared methods.} \label{nmi}
	\centering%
	\setlength{\tabcolsep}{1.5mm}{
		\begin{tabular}{|c|c|c|c|c|c|c|c|c|c|c|}%
			\hline
			~Datasets~   & ALLfea & LapScore  \cite{he2005laplacian} & UDFS  \cite{yang2011ℓ} & SOGFS  \cite{nie2016unsupervised} & RNE \cite{liu2020robust} & FSPCA  \cite{nie2022learning} & SPCAFS  \cite{li2021sparse} & SPCA-Net  \\ \hline\hline
			\multirow{2}{*}{COIL20} & 76.04$\pm$1.69 & 69.01$\pm$1.53 & 69.12$\pm$1.17 & 68.03$\pm$1.59 & 70.76$\pm$2.07 & 68.41$\pm$1.60 & 70.29$\pm$1.31 & \textbf{72.21$\pm$2.68} \\
			& (10) & (100) & (80) & (100) & (100) & (100) & (100) & \textbf{(90)} \\  \hline
			\multirow{2}{*}{Isolet} & 76.09$\pm$1.77 & 69.86$\pm$1.26 & 56.73$\pm$1.05 & 67.15$\pm$1.45 & 64.74$\pm$1.28 & 71.12$\pm$1.11 & 69.18$\pm$1.33 & \textbf{71.80$\pm$1.59} \\ 
			& (10) & (100) & (100) & (100) & (90) & (80) & (100) & \textbf{(100)} \\ \hline
			\multirow{2}{*}{UMIST} & 64.07$\pm$1.76 & 61.23$\pm$2.15 & 55.43$\pm$1.50 & 61.46$\pm$2.03 & 56.08$\pm$1.80 & 64.94$\pm$1.65 & 66.26$\pm$1.74 & \textbf{66.62$\pm$7.52} \\
			& (10) & (100) & (80) & (70) & (60) & (100) & (100) & \textbf{(70)} \\  \hline
			\multirow{2}{*}{MSTAR} & 83.96$\pm$3.14 & 73.90$\pm$1.62 & 78.18$\pm$3.64 & 76.56$\pm$1.54 & 78.26$\pm$2.51 & 78.87$\pm$2.52 & 79.62$\pm$2.30 & \textbf{80.67$\pm$3.47} \\ 
			& (10) & (100) & (90) & (100) & (100) & (90) & (100) & \textbf{(90)} \\
			\hline
		\end{tabular}
	}\label{pcanet_nmi}
\end{table*}

 \subsection{Ablation Studies}
Table \ref{net} lists the clustering results without and with the network, where $\times$ means that the regularization parameters are selected by grid search. It can be seen that deep unfolding networks can significantly improve ACC and NMI. In fact, In fact, it is quite difficult for grid search to find out the proper regularization parameters. In addition, Table \ref{dyn} lists the clustering results without and with dynamic parameters, where $\times$ indicates that the regularization parameters are fixed during the network iterations. Obviously, the dynamic parameters can better adapt to the data. The above ablation experiments validate the necessity of our method.

 		\begin{table}[t]
 	\renewcommand\arraystretch{1.15}
 		\caption{Ablation studies for the network.} \label{net}
 	\centering
	\setlength{\tabcolsep}{3.5mm}{
 		\begin{tabular}{|c|c|c|c|c|}
 			\hline
 			Datasets& Network &ACC & NMI  \\
 			\hline\hline
 			\multirow{2}*{COIL20}&
 			$\times$ & 55.12$\pm$2.67 & 70.44$\pm$1.37\\
 			& $\surd$ & \textbf{57.46$\pm$2.76 }&\textbf{72.21$\pm$2.68} \\
 			\hline
 			\multirow{2}*{Isolet}&
 			$\times$ & 51.84$\pm$2.82 & 67.02$\pm$1.43 \\
 			&$\surd$ &\textbf{58.43$\pm$4.31} & \textbf{71.80$\pm$1.59} \\
 			\hline
 			\multirow{2}*{UMIST}&
 			$\times$ & 40.65$\pm$2.29 &55.88$\pm$1.62 \\
 			&$\surd$ & \textbf{47.58$\pm$4.97} &\textbf{66.62$\pm$7.52} \\
 			\hline
 			\multirow{2}*{MSTAR}&
 			$\times$ & 80.65$\pm$6.47 &80.53$\pm$2.41 \\
 			&$\surd$ &\textbf{81.90$\pm$6.87} & \textbf{80.67$\pm$3.47} \\
 			\hline
 		\end{tabular}
 	}
 \end{table}

	\begin{table}[t]
	\renewcommand\arraystretch{1.15}
	\caption{Ablation studies for dynamic parameters.} \label{dyn}
	\centering
	\setlength{\tabcolsep}{3.5mm}{
		\begin{tabular}{|c|c|c|c|c|}
			\hline
			Datasets& Dynamic &ACC & NMI  \\
			\hline\hline
			\multirow{2}*{COIL20}&
			$\times$ & 56.71$\pm$3.83 & 71.49$\pm$3.67\\
			& $\surd$ & \textbf{57.46$\pm$2.76 }&\textbf{72.21$\pm$2.68} \\
			\hline
			\multirow{2}*{Isolet}&
			$\times$ & 52.06$\pm$3.71 & 68.91$\pm$2.36 \\
			&$\surd$ &\textbf{58.43$\pm$4.31} & \textbf{71.80$\pm$1.59} \\
			\hline
			\multirow{2}*{UMIST}&
			$\times$ & 42.63$\pm$2.78 &60.12$\pm$1.69 \\
			&$\surd$ & \textbf{47.58$\pm$4.97} &\textbf{66.62$\pm$7.52} \\
			\hline
			\multirow{2}*{MSTAR}&
			$\times$ & 80.74$\pm$5.28 &80.59$\pm$3.67 \\
			&$\surd$ &\textbf{81.90$\pm$6.87} & \textbf{80.67$\pm$3.47} \\
			\hline
		\end{tabular}
	}
\end{table}

\begin{figure}[t]
		\vspace{-37mm}
	\centering
	\includegraphics[width=0.55\textwidth]{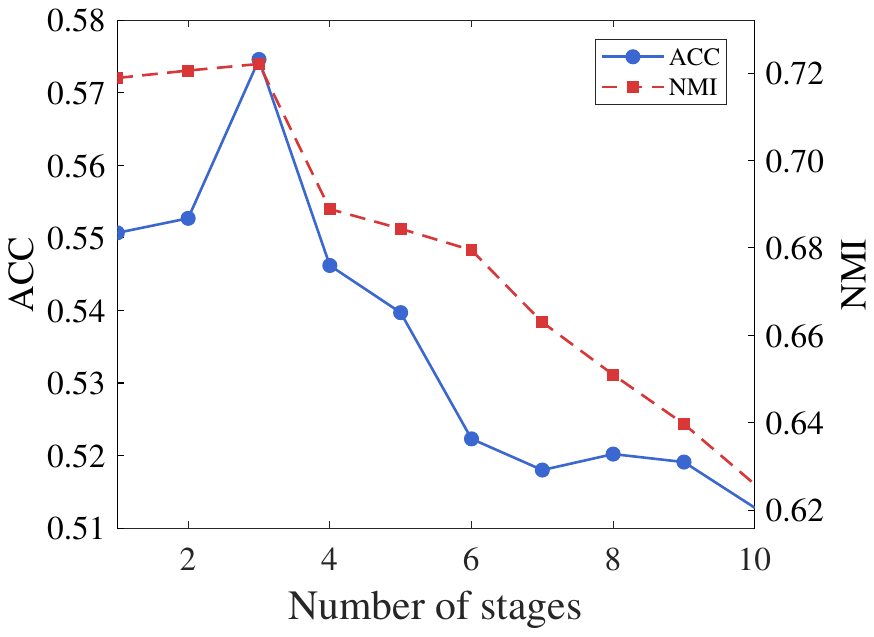}
		\vspace*{-42mm}
	\caption{Visualization of deep unfolding stages on COIL20.}
	\label{show}
\end{figure}

 \subsection{Remark}
Compared with traditional numerical optimization, which often requires hundreds or thousands of  iterations, deep unfolding networks only require less than 10 stages to achieve good results; see Fig. \ref{show}. However, it should be noted that the performance will degrade if the number of iterations exceeds a certain number, so in practice we chose between 3 and 5.

\section{Conclusion}

In this paper, we mathematically construct a structured sparse PCA model and combine it with deep unfolding networks to propose an encouraging UFS approach. It can not only learn the regularization parameters of the model, but also learn the penalty parameters in ADMM, thus providing a tuning-free algorithm. To the best of our knowledge, this is the first time that attempts to investigate deep unfolding networks for UFS. Experimental results demonstrate that even compared with the state-of-the-art FSPCA and SPCAFS, the proposed method still delivers superior performance in terms of ACC and NMI. 

In the future we are interested in designing more effective loss functions and extending it to deep equilibrium networks \cite{gilton2021deep} to further improve the performance.

\bibliographystyle{ieeetr}
\bibliography{mybibfile}

\end{document}